\documentclass{article}

\usepackage[preprint]{neurips_2026}

\usepackage{listings}
\usepackage{xcolor}
\usepackage[utf8]{inputenc} 
\usepackage[T1]{fontenc}    
\usepackage{hyperref}       
\usepackage{url}            
\usepackage{booktabs}       
\usepackage{amsfonts}       
\usepackage{nicefrac}       
\usepackage{microtype}      
\usepackage{xcolor}         
\usepackage{graphicx}
\usepackage{amsmath}
\usepackage{multirow}
\usepackage[table]{xcolor}
\usepackage{pifont}     %
\usepackage{subfig} 
\usepackage{enumitem}

\usepackage{CJKutf8}
\usepackage[utf8]{inputenc}
\usepackage{mdframed} 

\definecolor{citecolor}{HTML}{0071bc}
\definecolor{ourscolor}{HTML}{c2d1e5}

\title{Knowledge Visualization: A Benchmark and Method for Knowledge-Intensive Text-to-Image Generation}

\author{
    Ran Zhao\textsuperscript{\rm 1}, \quad
    Sheng Jin\textsuperscript{\rm 2},  \quad
    Size Wu\textsuperscript{\rm 3},  \quad
    Kang Liao\textsuperscript{\rm 3} \\
    \textbf{Zerui Gong\textsuperscript{\rm 3},  \quad
    Zujin Guo\textsuperscript{\rm 1},  \quad
    Yang Xiao\textsuperscript{\rm 1},  \quad
    Wei Li\textsuperscript{\rm 3}}
    \\
    \textsuperscript{\rm 1}Huazhong University of Science and Technology, 
    \textsuperscript{\rm 2}The University of Hong Kong \\
    \textsuperscript{\rm 3}S-Lab, Nanyang Technological University\\
    {\tt zhaoran@hust.edu.cn, size001@e.ntu.edu.sg}
}

\begin{document}

\maketitle

\begin{figure}[ht]
  \includegraphics[width=\textwidth]{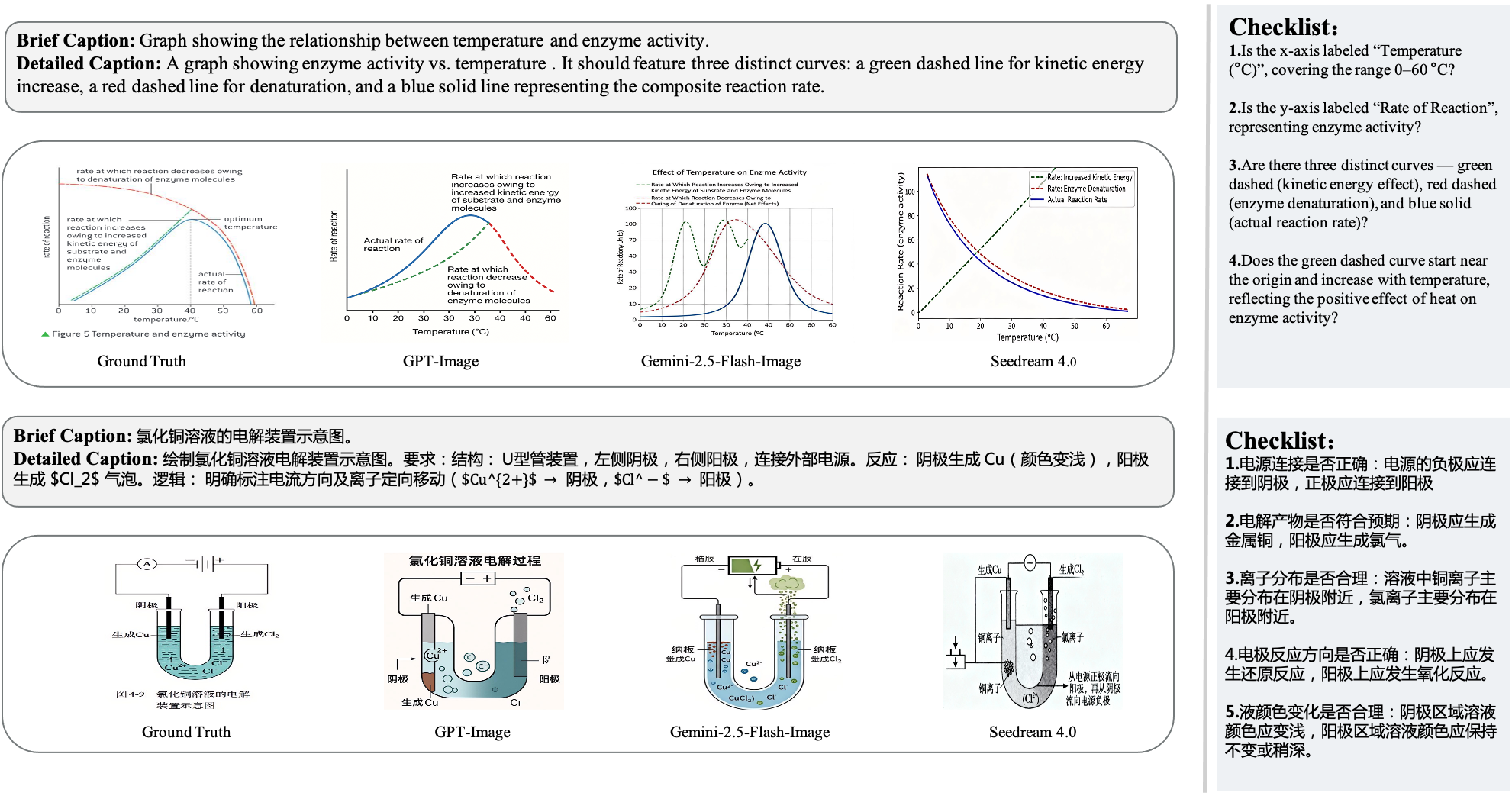}
  \caption{Illustrative examples from KVBench, featuring expert-curated prompts, textbook-grounded reference images, and a set of atomic checklist-based evaluation criteria.}
  \label{fig:sample}
  \vspace{3mm}
\end{figure}

\begin{abstract}
Recent text-to-image (T2I) models have demonstrated impressive capabilities in photorealistic synthesis and instruction following. However, their reliability in knowledge-intensive settings remains largely unexplored. Unlike natural image generation, knowledge visualization requires not only semantic alignment but also strict adherence to domain knowledge, structural constraints, and symbolic conventions, exposing a critical gap between visual plausibility and scientific correctness.
To systematically study this problem, we introduce KVBench, a curriculum-grounded benchmark for evaluating knowledge-intensive T2I generation. KVBench covers six senior high-school subjects: Biology, Chemistry,  Geography, History, Mathematics, and Physics. The benchmark consists of 1,800 expert-curated prompts derived from over 30 authoritative textbooks. Each sample is paired with a reference image and a fine-grained evaluation checklist that decomposes correctness into atomic and verifiable constraints, enabling objective, interpretable, and reproducible assessment of structural accuracy and reasoning consistency.
Using this benchmark, we evaluate 14 state-of-the-art open- and closed-source models, revealing substantial deficiencies in logical reasoning, symbolic precision, and multilingual robustness, with open-source models consistently underperforming proprietary systems.
To address these limitations, we further propose KE-Check, a two-stage framework that improves scientific fidelity via (1) Knowledge Elaboration for structured prompt enrichment, and (2) Checklist-Guided Refinement for explicit constraint enforcement through violation identification and constraint-guided editing. KE-Check effectively mitigates scientific hallucinations, narrowing the performance gap between open-source and leading closed-source models. Data and codes are publicly available at \url{https://github.com/zhaoran66/KVBench}. 
\end{abstract}

\section{Introduction}

Recent advances in text-to-image (T2I) generation have achieved remarkable progress in photorealism and instruction following~\cite{rombach2022high,betker2023improving,esser2024scaling,xiao2025omnigen,wu2025qwenimagetechnicalreport,flux2024,flux-2-2025}. Modern generative models can synthesize visually compelling images from open-ended textual descriptions, demonstrating strong capabilities in composition, style control, and semantic alignment. 

However, as these models are increasingly deployed in knowledge-intensive scenarios such as scientific education, textbook illustration, and academic communication, a critical limitation emerges: \emph{visually plausible generation does not imply scientific correctness}~\cite{lu2022learn,fu2025evaluating}. Existing T2I systems frequently produce images that appear reasonable at a glance, yet violate fundamental domain knowledge, including incorrect structures, missing components, inconsistent spatial relations, and erroneous symbolic representations.

This limitation stems from a fundamental mismatch between natural image synthesis and knowledge visualization. While natural image generation prioritizes perceptual realism and coarse semantic alignment, knowledge visualization demands \emph{precise compliance with domain-specific constraints}, including structural correctness, symbolic accuracy, and standardized visual conventions. Generating such images requires not only understanding textual descriptions, but also grounding them in structured knowledge and enforcing fine-grained constraints during generation.

Despite its importance, this problem remains underexplored. Recent efforts have begun to evaluate T2I models using exam-style or multidisciplinary benchmarks~\cite{luo2025mmmg,chang2025b,wang2025genexam}, focusing on broad reasoning capabilities across diverse domains. However, these settings are loosely defined and often rely on holistic judgment, making it difficult to assess whether generated images satisfy the precise requirements of real-world educational content. In practice, educational applications require \emph{curriculum-grounded visualization}, where correctness is defined by strict textbook standards rather than subjective plausibility.

To bridge this gap, we introduce \textbf{KVBench}, a curriculum-grounded benchmark for knowledge-intensive text-to-image generation. KVBench focuses on knowledge visualization tasks derived from senior high-school education, covering six core subjects: Biology, Chemistry, Geography, History, Mathematics, and Physics. Each sample  requires generating a textbook-quality image that satisfies rigorous domain constraints. The benchmark contains 1,800 expert-curated bilingual samples, enabling systematic evaluation of both reasoning accuracy and multilingual robustness.

A key feature of KVBench is its checklist-based verification pipeline. Instead of relying on holistic scoring, we decompose each sample into a set of atomic, verifiable checklist items derived from authoritative textbooks. These items explicitly encode requirements on entities, spatial relations, labels, and visual conventions, enabling fine-grained and interpretable evaluation of model outputs. All samples and checklists are curated from over 30 textbooks and validated through multi-stage expert review, ensuring pedagogical fidelity and reproducibility.

We further propose \textbf{KE-Check}, a simple yet effective framework for improving scientific fidelity. KE-Check follows a two-stage design. First, \emph{Knowledge Elaboration} enriches abstract prompts into structured, domain-specific descriptions, providing a stronger conditioning signal for generation. Second, \emph{Checklist-Guided Refinement} explicitly enforces constraint satisfaction by constructing structured checklists, performing item-wise constraint auditing, and applying constraint-guided editing. By decoupling knowledge injection from constraint enforcement, KE-Check effectively mitigates scientific hallucinations and produces high-fidelity visualizations. Notably, it significantly narrows the performance gap between open-source and proprietary models.

In summary, this work makes the following contributions:
\begin{itemize}

\item \textbf{KVBench: A Curriculum-Grounded Benchmark for Knowledge Visualization.} 
We introduce KVBench, a standardized benchmark for evaluating knowledge-intensive T2I generation. The dataset consists of 1,800 bilingual prompts curated from over 30 authoritative textbooks across six subjects. We further propose a checklist-based verification pipeline that decomposes visualization correctness into atomic, verifiable checklist items, enabling objective, interpretable, and reproducible assessment beyond holistic metrics.

\item \textbf{KE-Check: A Constraint-Aware Refinement Framework.}
We propose KE-Check, a two-stage framework that combines Knowledge Elaboration with Checklist-Guided Refinement to enforce scientific correctness in generated images. Our method explicitly models constraint violations and performs constraint-guided editing, substantially improving structural accuracy and symbolic fidelity.

\item \textbf{Comprehensive Analysis of Knowledge-Intensive T2I.}
We conduct extensive experiments on state-of-the-art T2I models, revealing persistent deficiencies in reasoning, constraint satisfaction, and multilingual robustness. Our findings highlight key challenges in knowledge-grounded generation and provide actionable insights for future research.

\end{itemize}

\section{Related Works}

\subsection{Text-to-Image Generation Models.}
Text-to-image (T2I) generation has progressed rapidly from pixel-space diffusion models~\cite{ho2020denoising, dhariwal2021diffusion} to high-fidelity latent approaches such as Stable Diffusion~\cite{rombach2022high}, the Flux series~\cite{bfl2025flux2, labs2024flux}, and DALL·E~\cite{ramesh2021zero, betker2023improving}. These models achieve impressive visual quality and semantic alignment, but are primarily optimized for perceptual realism rather than structured correctness. Recent advances have shifted toward autoregressive and unified multimodal architectures, which model images and text within a shared token space~\cite{sun2024autoregressive, li2024seed, tian2024visual}. Unified Multimodal Large Language Models (MLLMs), including Chameleon~\cite{chameleon2024}, Emu3~\cite{wang2024emu3}, and subsequent frameworks~\cite{xie2025unified, deng2025unified}, demonstrate strong capabilities in both generation and understanding by treating visual content as sequential representations.

More recently, reasoning-driven generation has emerged as a new frontier, aiming to handle complex and rule-constrained tasks. Models such as GPT-Image-1~\cite{openai2025b}, Gemini 2.5-Flash-Image~\cite{google2025}, and Gemini-3-pro-image~\cite{gemini3report2026} incorporate implicit or explicit reasoning processes for visual synthesis. Complementary approaches, including Chain-of-Thought prompting for image generation~\cite{li2025thought} and step-wise reasoning paradigms~\cite{guo2025reasoning,lian2023llm}, further enhance compositional control. Despite these advances, existing models remain limited in producing scientifically rigorous visualizations, where correctness is governed by strict structural, relational, and symbolic constraints. This gap motivates the need for both specialized evaluation and constraint-aware generation frameworks.

\subsection{Text-to-Image Evaluation Benchmarks.}
Evaluation of T2I systems has evolved from holistic distribution-based metrics, such as FID~\cite{heusel2017gans} and CLIPScore~\cite{hessel2021clipscore}, to fine-grained alignment and reasoning-oriented benchmarks. Early works like GenEval~\cite{ghosh2023geneval} and TIFA~\cite{hu2023tifa} focus on compositional correctness with explicitly defined visual elements. Subsequent benchmarks increase prompt complexity and diversity~\cite{wei2025bench, zhou2025bench, fang2025bench}, but often fall short in capturing the dense constraints present in real-world applications. To better assess higher-level cognition, reasoning-centric benchmarks have been proposed, evaluating dimensions such as commonsense reasoning, logical consistency, and causal understanding~\cite{fu2024mme, meng2024reasoning, niu2025evaluating, gong2025reasoning}. While these benchmarks advance general-domain evaluation, they largely overlook the strict requirements of knowledge visualization. Several works have begun exploring academic scenarios, including MMMG~\cite{luo2025mmmg}, OneIG-Bench~\cite{chang2025a}, and SridBench~\cite{chang2025b}. However, they rely on holistic scoring that lacks fine-grained diagnostic capability. In contrast, our proposed KVBench is explicitly designed for curriculum-grounded knowledge visualization. It differs in three key aspects: (1) \emph{textbook-aligned knowledge visualization tasks} with domain constraints; (2) a \emph{checklist-based evaluation protocol} that decomposes correctness into atomic, verifiable criteria; and (3) \emph{bilingual prompt design} for evaluating multilingual robustness. These design choices enable precise, interpretable, and reproducible evaluation, addressing a critical gap in current T2I benchmarking.

\section{KVBench}

\subsection{Overview}

We introduce \textbf{KVBench}, a new benchmark meticulously designed to evaluate the compositional accuracy and scientific reasoning capabilities of text-to-image (T2I) models in knowledge-intensive generation. Unlike general-purpose T2I benchmarks, KVBench is distinguished by prompts demanding high compositional fidelity and complex scientific reasoning, reflecting the rigor required for knowledge visualization.

KVBench systematically covers six core high-school academic subjects: Biology, Chemistry, Geography, History, Mathematics, and Physics. For each subject, we curated 150 unique knowledge concepts, with prompts provided in both English and Chinese. This results in a total of 1,800 evaluation samples, ensuring broad and diverse coverage.

\begin{figure*}[t]
\centering
\subfloat{
  \includegraphics[width=0.45\linewidth]{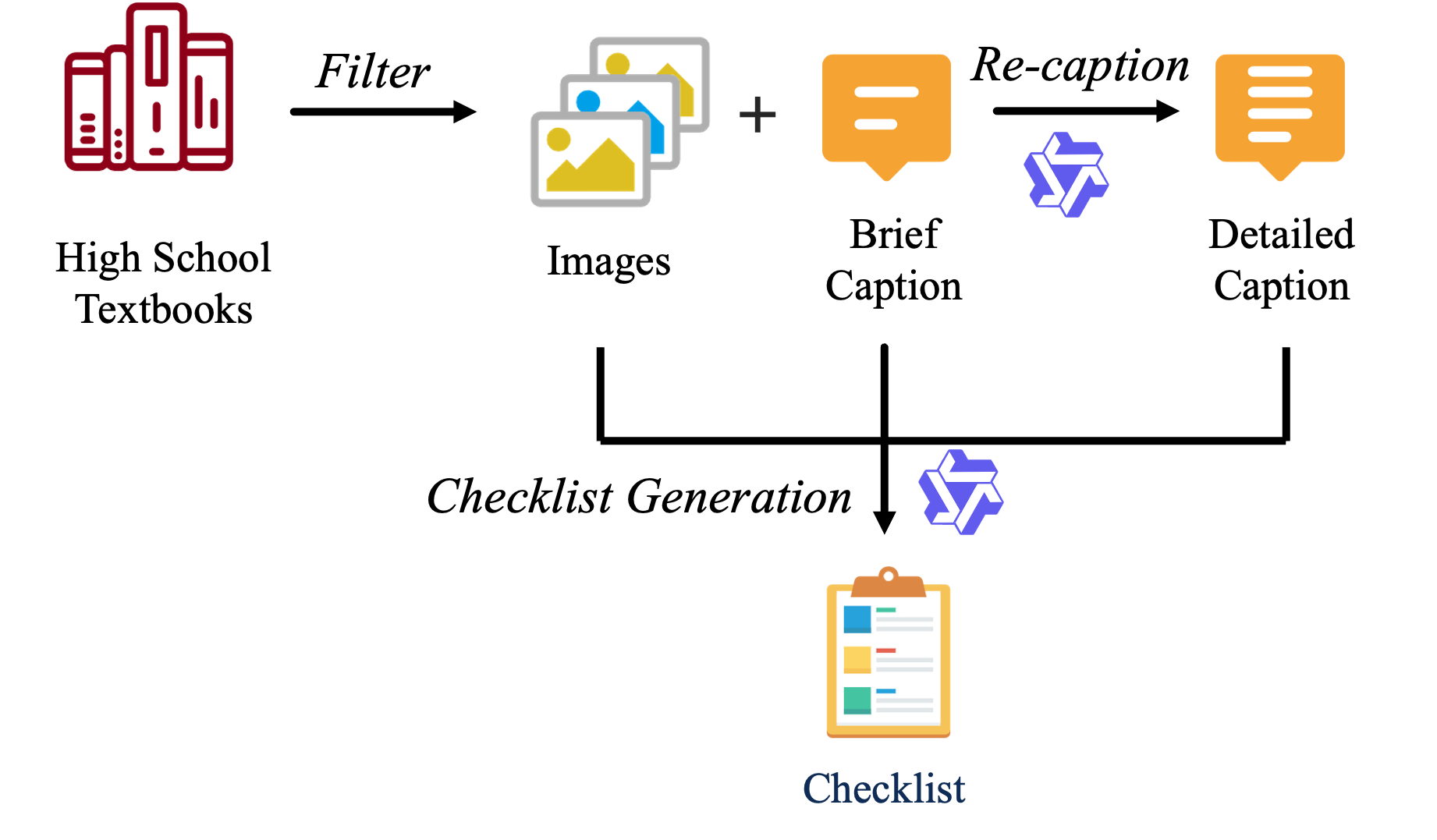}
}
\hfill
\subfloat{
  \includegraphics[width=0.45\linewidth]{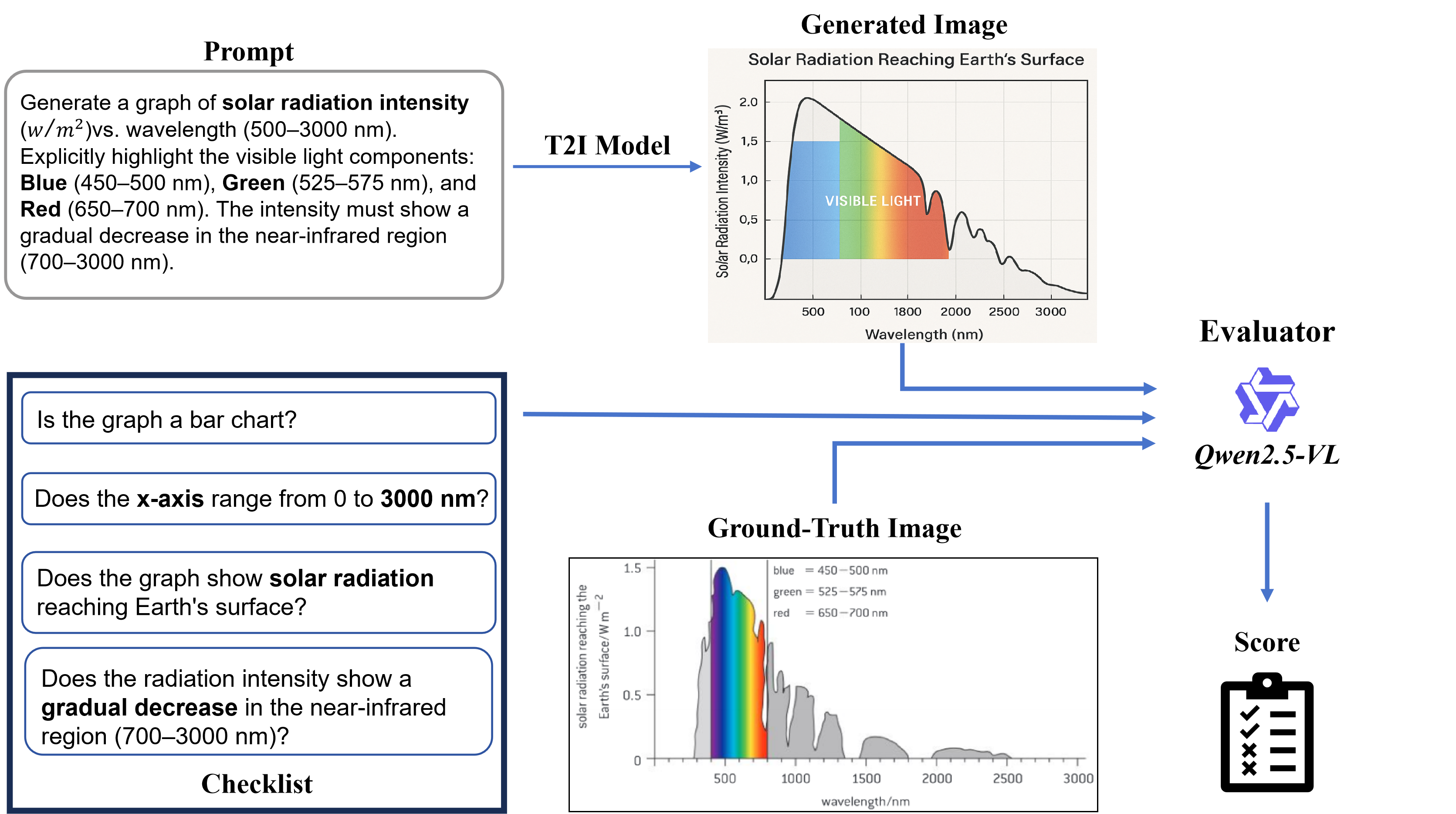}
}
\caption{Data construction pipeline (left) and evaluation pipeline (right) of KVBench.}
\label{fig:data_construction}
\end{figure*}

\subsection{Dataset Construction}

Our construction process, depicted in Figure~\ref{fig:data_construction}, prioritizes scientific accuracy, educational relevance, and verifiable ground truth.

\textit{Data Collection.}
To ensure a systematic and standardized knowledge base, KVBench is built upon senior high school curricula. In contrast to benchmarks that draw from web-crawled data, which may contain factual errors or generation artifacts~\cite{gehrmann2023repairing,ji2023survey}, KVBench is grounded exclusively in authoritative textbooks. Our collection process begins by using Multimodal Large Language Models (MLLMs) to pre-filter textbook content, identifying schematic diagrams and illustrations while extracting their original captions. Subsequently, a team of PhD-level domain experts manually curates this collection. This expert-led curation ensures each selected image is scientifically accurate, visually clear, and conceptually rich, while its corresponding caption provides sufficient detail for image generation. This methodology guarantees curriculum alignment and freedom from web-based contamination, making the benchmark both reliable and educationally meaningful.

\textit{Prompt Generation.}
For each reference image, we generate a pair of prompts to assess different model capabilities. These prompts consist of a \textit{Brief Caption} and a \textit{Detailed Caption}. The brief caption, extracted directly from the textbook, captures the high-level concept. To generate the detailed caption, we employ a re-captioning technique~\cite{betker2023improving} using a powerful MLLM (Qwen2.5-VL-32B), which takes the reference image and its brief caption as input to produce a comprehensive description of all fine-grained visual elements. Crucially, these MLLM-generated detailed captions are then manually reviewed and refined by a team of five PhD-level experts to ensure semantic correctness, knowledge density, and robustness against trivial formatting changes.

\subsection{Evaluation Method}
Evaluating knowledge-intensive T2I generation demands a level of granularity that conventional metrics cannot provide. Metrics like CLIPScore~\cite{hessel2021clipscore} struggle to resolve multiple fine-grained elements, while holistic MLLM-based scoring~\cite{lin2024evaluating} is susceptible to hallucinations and error propagation~\cite{zheng2023judging}.

To overcome these limitations, we propose a systematic, checklist-based verification pipeline inspired by~\cite{hu2023tifa}. Each task in KVBench is associated with an atomic checklist that decomposes the complex scientific concept into a set of orthogonal, binary questions~\cite{hu2023tifa,lee2025checkeval}. These checklist items cover: (1) the presence of key object instances, (2) correctness of visual attributes, (3) fidelity of spatial relationships~\cite{huang2023t2i}, and (4) accuracy of reasoning-derived outcomes~\cite{lu2023mathvista}. The checklists are generated by an MLLM (Qwen2.5-VL-32B) conditioned on the reference image and both captions, and are grounded in textbook facts to ensure pedagogical correctness~\cite{suris2023vipergpt}.

Evaluation proceeds by framing each checklist item as a binary visual question-answering task. An MLLM evaluator assigns a score \(S_i \in \{0, 1\}\) for the \(i\)-th item, where \(1\) denotes success ("Yes") and \(0\) denotes failure ("No"). The final score for a sample is the mean accuracy across all \(N\) items, yielding a fine-grained alignment score: $S = \frac{1}{N} \sum_{i=1}^{N} S_i$.

This approach grounds evaluation in verifiable facts rather than subjective preferences~\cite{lee2025checkeval}. For scalable and reproducible assessment, we employ Qwen2.5-VL-32B-Instruct as our primary MLLM evaluator, chosen for its state-of-the-art vision-language capabilities. Each sample is paired with a diagnostic checklist typically consisting of four to six items. In total, the benchmark comprises 1,800 sample instances and 5,158 corresponding checklist items.

\textit{Evaluation Protocol.}
To disentangle a model's internalized knowledge from its instruction-following abilities, we establish two distinct evaluation tracks. In the \textbf{Brief Caption} track, models must rely on implicit knowledge and reasoning to reconstruct scientific concepts from abstract prompts. Conversely, the \textbf{Detailed Caption} track assesses the model's capacity to faithfully render complex visual information from explicit, fine-grained constraints. This dual-track design enables a more granular diagnosis of model failures in knowledge-intensive T2I tasks~\cite{cho2023visual,lian2023llm,qu2023layoutllm}.

\subsection{Human Alignment}
\label{sec:human_alignment}

To validate our automated protocol, we conducted a rigorous human-AI agreement study. As detailed in Table~\ref{tab:human_ai}, our automatic evaluation pipeline, using Qwen2.5-VL, achieves a high degree of alignment with human expert annotations. Specifically, it reaches an overall accuracy of 80.26\% and a substantial Cohen's kappa of \(\kappa=0.7447\) (\(p<0.0001\)). This high level of consistency, especially given the specialized nature of the academic content, validates our checklist-based framework as a reliable and objective proxy for human judgment in evaluating knowledge-intensive T2I generation~\cite{lin2024evaluating,zheng2023judging}.

\begin{table}[t]
    \centering
    \small
    \caption{Human-AI alignment accuracy: Comparison of MLLM evaluators across English and Chinese datasets against human expert ground truth.}
    \label{tab:human_ai}
    \begin{tabular}{lccc}
        \toprule
        \textbf{Model} & \textbf{English (Acc)} & \textbf{Chinese (Acc)} & \textbf{Overall (Mean)} \\
        \midrule
        \textbf{GPT-4o}       & \textbf{0.8433} & 0.7778 & \textbf{0.8106} \\
        \textbf{Gemini-2.0}   & 0.6833 & \textbf{0.8000} & 0.7417 \\
        \textbf{Qwen2.5-VL}   & 0.8095 & 0.7956 & 0.8026 \\
        \bottomrule
    \end{tabular}
\end{table}

\begin{figure}[t]
  \centering
  \includegraphics[width=0.99\linewidth]{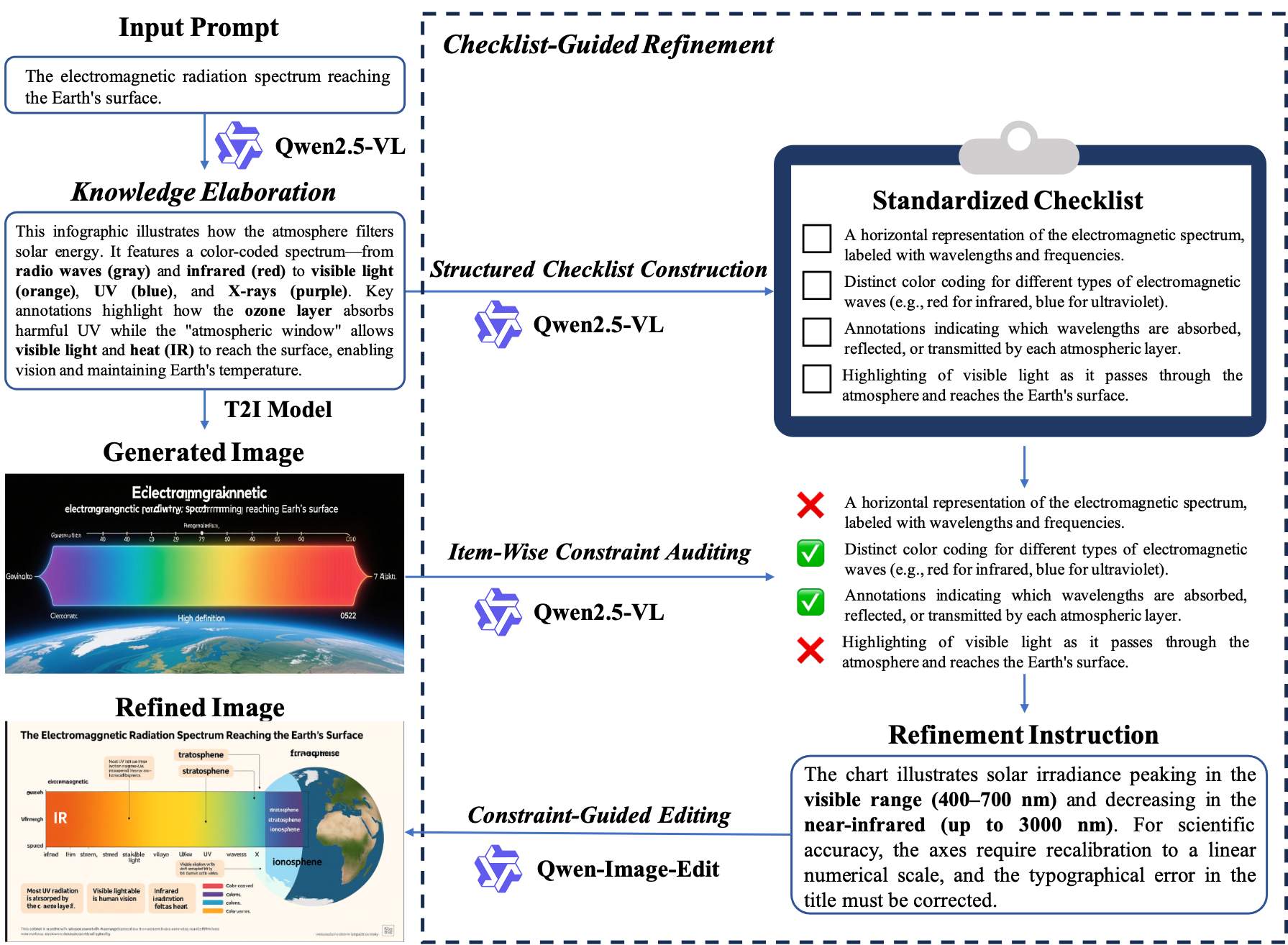}
  \caption{Overview of KE-Check. Knowledge Elaboration expands the input prompt into a domain-specific description for image generation. Checklist-Guided Refinement constructs a structured checklist, performs item-wise auditing to identify violations, and applies constraint-guided editing.}
  \label{fig:method}
\end{figure}

\section{Method}
We propose \textbf{KE-Check}, a two-stage framework for high-fidelity scientific knowledge visualization from textual prompts. Unlike generic text-to-image (T2I) synthesis, which primarily emphasizes perceptual realism, knowledge visualization requires strict adherence to domain knowledge, structural logic, and standardized visual conventions. KE-Check addresses this challenge through two sequential components: \textbf{Knowledge Elaboration} and \textbf{Checklist-Guided Refinement}.

Rather than relying on ambiguous holistic generation, our framework first enriches brief prompts with domain-specific semantics to produce a structurally grounded initial image. It then performs targeted, constraint-aware refinement via a structured checklist. This design explicitly decouples knowledge-enhanced generation from error diagnosis and correction, prioritizing scientific fidelity over superficial plausibility.

\subsection{Problem Formulation}

Knowledge-intensive image generation fundamentally differs from natural image synthesis. Given a concise textual prompt $x$ describing a scientific concept, a T2I model produces an image $I = G(x)$. In this setting, the quality of the generated visualization is determined not by photorealism, but by its compliance with domain-specific structural, relational, and symbolic constraints.

We formalize the requirements of a valid knowledge visualization as a set of atomic constraints:
\begin{equation}
\mathcal{C}(x) = \{ c_1, c_2, \dots, c_N \},
\end{equation}
where each constraint $c_i$ represents a verifiable rule over entities, spatial relations, labels, attributes, or domain conventions derived from scientific knowledge.

Under this formulation, scientific hallucinations can be precisely characterized as constraint violations:
\begin{equation}
\mathcal{V}(I, \mathcal{C}) = \{ c_i \mid I \not\models c_i \}.
\end{equation}

Consequently, the refinement can be viewed as a constraint-grounded editing process: identifying violated constraints and transforming them into actionable guidance for faithful image editing.

\subsection{Knowledge Elaboration}

Directly generating visualizations from concise prompts results in oversimplified content, missing details, and severe hallucinations. This limitation arises because concise prompts typically encode only high-level semantics, lacking the domain knowledge and fine-grained visual specifications required for accurate visualization.

To address this issue, we introduce Knowledge Elaboration, a prompt enhancement module tailored for knowledge visualization. Given a concise input prompt, this module expands it into a comprehensive, structured description that integrates two key aspects. (1) Domain knowledge: including core concepts, component hierarchies, spatial organization rules, and domain-specific conventions. (2) Visual specifications: including graphical primitives, line styles, color schemes, and layout details aligned with scientific visualization standards.

The resulting knowledge-rich prompt provides a strong conditioning signal for the T2I model, enabling it to generate an image $I$ with improved structural completeness and scientific consistency.

\subsection{Checklist-Guided Refinement}

Despite improved initial generation, one-pass synthesis remains insufficient to fully eliminate scientific inaccuracies. To enforce strict compliance with domain constraints, we propose Checklist-Guided Refinement, a structured pipeline consisting of three steps: \textit{checklist construction}, \textit{constraint auditing}, and \textit{constraint-guided editing}.

\subsubsection{Structured Checklist Construction}

We first construct a structured evaluation checklist that formalizes scientific requirements into explicit, verifiable atomic constraints~\cite{hu2023tifa,lee2025checkeval}. Given the elaborated knowledge description, a multimodal large language model (MLLM) generates a standardized checklist covering key evaluation dimensions, including scientific correctness, structural integrity, detail completeness, label accuracy, and visual standardization.

Each checklist item corresponds to an independent, atomic constraint, transforming implicit quality expectations into explicit and measurable criteria. This checklist serves as a unified reference for subsequent auditing and refinement.

\subsubsection{Item-Wise Constraint Auditing}

We then perform fine-grained, item-wise auditing~\cite{yang2023dawn, yu2023mm,suris2023vipergpt,cho2023visual} to verify whether the generated image satisfies each constraint. The violation of each constraint is defined as:
\begin{equation}
v_i =
\begin{cases}
1 & \text{if } I \not\models c_i \\
0 & \text{otherwise}
\end{cases}
\end{equation}

Unlike holistic evaluation, this step examines each constraint individually, enabling precise identification of errors such as missing components, incorrect spatial relations, mislabeled elements, and non-standard visual representations. The resulting violation set $\mathcal{V}$ provides interpretable and localized feedback for refinement.

\subsubsection{Constraint-Guided Editing}

Naively correcting each violation independently may lead to fragmented or conflicting edits, potentially disrupting global consistency. To address this, we aggregate the detected violations into coherent, non-redundant refinement instructions. Based on this consolidated guidance, we perform constraint-guided editing to correct only the non-compliant regions of the image, while preserving structurally valid and semantically correct content. This targeted refinement avoids unnecessary modifications, preventing semantic drift and visual degradation. The final output $I^*$ is a scientifically accurate and standardized visualization that satisfies all constraints.

\begin{table*}[t]
\centering
\caption{Model performance across 12 subjects (Brief Caption). \textbf{Overall} represents the average capability across all six academic subjects for each model. \textbf{Bold} values indicate best performance.}
\label{tab:subject2_with_overall_calculated}\resizebox{\textwidth}{!}{\begin{tabular}{l|cc|cc|cc|cc|cc|cc|cc}\toprule\multirow{2}{*}{\textbf{Model}} &\multicolumn{2}{c|}{\textbf{Overall}} &\multicolumn{2}{c|}{\textbf{Biology}} &\multicolumn{2}{c|}{\textbf{Chemistry}} &\multicolumn{2}{c|}{\textbf{Geography}} &\multicolumn{2}{c|}{\textbf{History}} &\multicolumn{2}{c|}{\textbf{Math}} &\multicolumn{2}{c}{\textbf{Physics}} \\
 & \textbf{zh} & \textbf{en} 
 & \textbf{zh} & \textbf{en} 
 & \textbf{zh} & \textbf{en} 
 & \textbf{zh} & \textbf{en} 
 & \textbf{zh} & \textbf{en} 
 & \textbf{zh} & \textbf{en} 
 & \textbf{zh} & \textbf{en} \\\midrule\multicolumn{15}{c}{\textit{Close-source Models}} \\\midrule
FLUX.2-max\cite{bfl2025flux2}        & 21.97 & 34.98 & 22.80 & 43.29 & 33.42 & 45.20 & 31.96 & 18.71 & 10.75 & 33.07 & 24.89 & 34.60 & 27.97 & 35.07 \\
GPT-Image\cite{betker2023improving}         & 40.90 & \textbf{41.60} & 38.00 & 46.00 & 50.50 & 51.60 & 44.31 & \textbf{45.96} & 23.65 & 25.73 & 29.33 & 38.04 & 39.10 & 42.20 \\
Seedream-4.0\cite{seedream2025seedream}      & 44.74 & 39.61 & 44.98 & 44.44 & 54.35 & 49.00 & 52.29 & 37.87 & \textbf{35.92} & 26.47 & 35.60 & 40.38 & 40.27 & 39.53 \\
gemini-3-pro-image\cite{gemini3report2026} & \textbf{51.36} & 34.40 & \textbf{69.35} &\textbf{57.24}  & \textbf{70.67} & 50.40 & \textbf{63.29} & 14.00 & 10.17 & 32.87 & \textbf{41.29} & 37.07 & \textbf{51.58} & 37.20 \\\midrule\multicolumn{15}{c}{\textit{Open-source Models}} \\\midrule
Janus-Pro\cite{chen2025janus}         & 8.81  & 21.61 & 6.64  & 27.71 & 13.35 & 27.07 & 12.53 & 18.49 & 4.83 & 6.67 & 10.49 & 23.38 & 5.08 & 26.33 \\
FLUX.1-Krea-dev\cite{flux1kreadev2025}   & 11.45 & 18.19 & 7.93 & 18.13 & 17.30 & 18.40 & 10.67 & 18.58 & 1.17 & 7.67 & 21.02 & 27.58 & 10.57 & 18.73 \\  
SD3.5-Large\cite{calcuis2024sd35gguf}    & 12.18 & 31.07 & 10.09 & 29.80 & 14.22 & 36.53 & 17.78 & 38.44 & 4.50 & 24.07 & 20.49 & 28.84 & 6.00 & 28.73 \\
show-o2\cite{xie2025showo2}           & 14.12 & 19.05 & 5.93 & 17.87 & 15.62 & 26.67 & 22.71 & 15.11 & 6.67 & 4.67 & 23.56 & 26.22 & 13.33 & 23.73 \\
BAGEL-7B-MoT~\cite{deng2025bagel}     & 18.75 & 22.20 & 14.13 & 28.64 & 18.73 & 23.73 & 27.91 & 19.07 & 14.83 & 10.13 & 14.09 & 21.98 & 22.78 & 29.67 \\
Z-Image\cite{team2025zimage}           & 19.15 & 19.14 & 13.11 & 16.49 & 21.25 & 23.67 & 27.33 & 16.04 & 19.97 & 13.00 & 19.82 & 25.64 & 13.43 & 20.00 \\
FLUX.1-dev~\cite{flux2024}       & 21.47 & 33.93 & 10.91 & 28.00 & 21.82 & \textbf{41.13} & 25.51 & 22.49 & 22.83 & 26.33 & 20.93 & 34.22 & 26.80 & \textbf{50.40} \\
OmniGen2\cite{wu2025omnigen2}          & 23.79 & 18.47 & 17.76 & 15.16 & 18.25 & 21.80 & 27.13 & 18.53 & 16.42 & 9.67 & 36.31 & 24.62 & 26.87 & 26.07 \\
FLUX.2-dev\cite{bfl2025flux2}        & 40.51 & \textbf{36.88} & 42.82 & 31.64 & 37.00 & 38.73 & 49.62 & 41.56 & 39.92 & \textbf{35.80} & 28.27 & \textbf{35.42} & 36.40 & 39.13 \\ \midrule
Qwen-Image\cite{wu2025qwenimagetechnicalreport}        & 20.09 & 25.74 & 14.76 & 20.56 & 15.17 & 23.53 & 29.00 & \textbf{27.87} & 21.97 & 27.33 & 19.16 & 27.16 & 20.53 & 28.00 \\
\rowcolor{ourscolor}
Qwen-Image + KE-Check  (Ours)                       & \textbf{47.67} & 30.24 & \textbf{54.73} & \textbf{36.05} & \textbf{55.47} & 34.93 & \textbf{55.87} & 27.82 & \textbf{38.75} & 18.00 & \textbf{34.89} & 33.15 & \textbf{46.33} & 31.47 \\\bottomrule\end{tabular}}
\end{table*}

\section{Experiments}
\subsection{Evaluated Models}

To benchmark the performance of state-of-the-art models on KVBench, we evaluate a diverse suite of 14 cutting-edge models. These models represent the current state-of-the-art in text-to-image (T2I) generation and multimodal understanding as of early 2026, spanning diverse architectural paradigms including Diffusion Transformers (DiT)~\cite{flux1kreadev2025,flux2024,bfl2025flux2,rombach2022high,yu2022scaling}, Autoregressive (AR) models~\cite{ma2025janusflow,xie2025showo2,cai2025z}, and Unified Multimodal Models (UMM)~\cite{wu2025qwen,xiao2025omnigen}.

\subsection{Benchmark Results}
Table~\ref{tab:subject2_with_overall_calculated} presents a comprehensive comparison of model performance across six academic disciplines and two languages (``zh'' for Chinese and ``en'' for English).

\textit{Closed-source models establish a substantial performance margin.}
Proprietary foundation models maintain clear advantages over open-weight alternatives across most subjects and linguistic settings.
Such performance gaps are especially salient in reasoning-heavy fields including Mathematics and Physics, where accurate symbolic expression and multi-step logical reasoning are essential.
Although several state-of-the-art open models achieve competitive results on individual subjects, none exhibits consistent and robust performance across all disciplines.

\textit{Knowledge visualization remains linguistically fragile.}
Models generally achieve higher scores under English prompts than Chinese ones.
This cross-lingual discrepancy reveals inherent limitations in multilingual knowledge rendering, academic term grounding, and structured character layout generation in modern text-to-image systems~\cite{cai2024benchlmm, liu2024mmbench}.

\textit{Discipline-Level Observations.}
Further inspection of per-subject performance unveils pronounced disparities in model capabilities across scientific and humanities disciplines.
Biology and Chemistry consistently yield superior generation quality across most evaluated models, primarily attributed to standardized formulaic representations and abundant domain-specific instructional data.
Conversely, disciplines such as History and Geography exhibit inferior performance, due to their heightened sensitivity to nuanced linguistic expressions and culturally grounded contextual knowledge.

\begin{figure*}[t]
  \centering
  \includegraphics[width=0.99\linewidth]{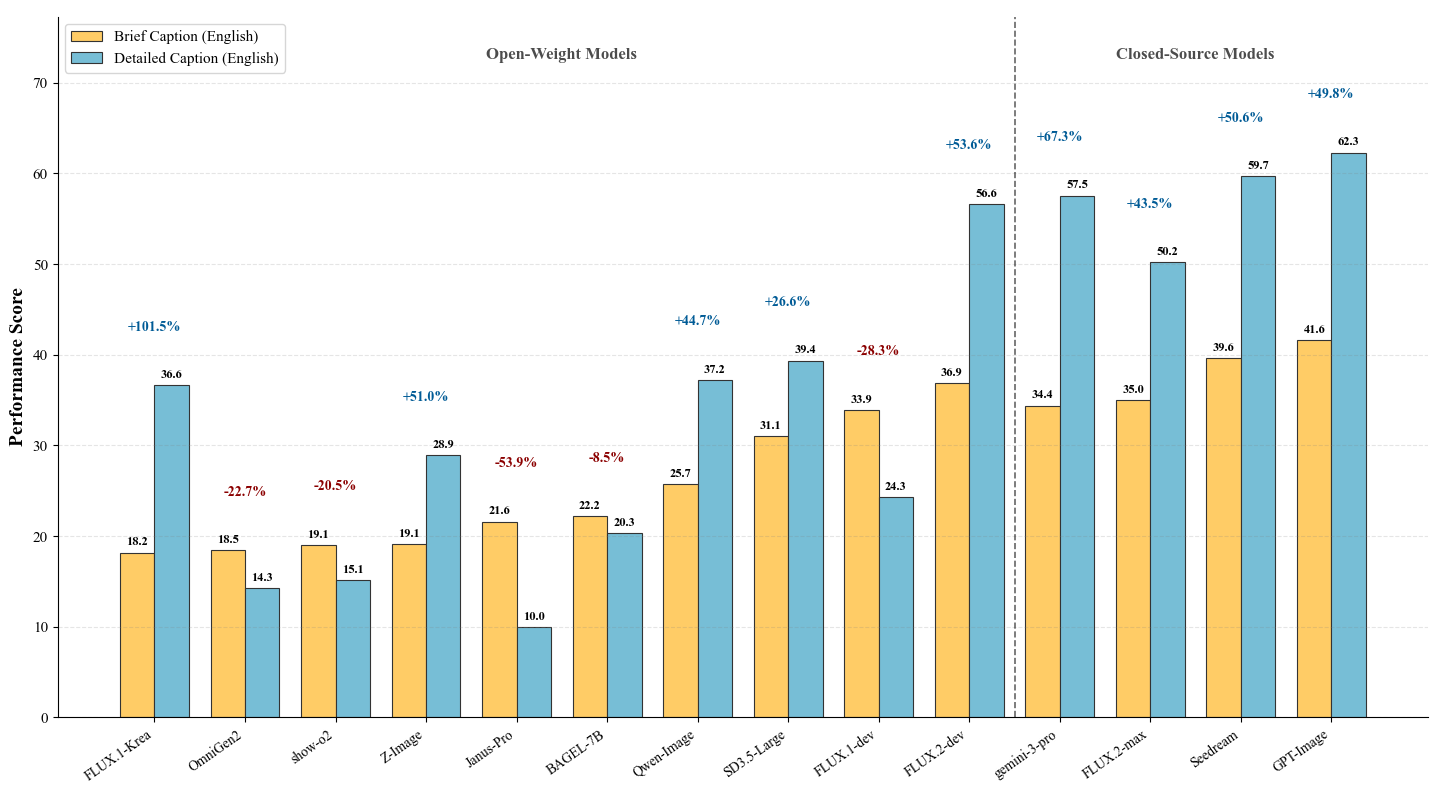}
  \caption{Performance comparison of open-weight and closed-source models under two prompt settings: Brief Caption and Detailed Caption. The percentage labels indicate the relative performance change of Detailed Caption compared to Brief Caption.}
  \label{fig:result_group}
\end{figure*}

\textit{Brief Caption v.s. Detailed Caption.}
As shown in Fig.~\ref{fig:result_group}, feeding Detailed Caption as the input prompt yields better model performance than using Brief Caption in most cases. Nevertheless, we unexpectedly observe a performance degradation when Detailed Caption is adopted for several weaker open-source models, including Janus-Pro~\cite{ma2025janusflow} and OmniGen2~\cite{xiao2025omnigen}. This counterintuitive result can be attributed to the inferior instruction-following capabilities of these models. Even with detailed descriptive prompts, such models fail to faithfully render complex visual layouts as required. Instead, redundant and detailed textual descriptions introduce extra interference to the generation process, which ultimately impairs the final output quality rather than enhancing it.

\subsection{Ablation Study of KE-Check}
Table~\ref{tab:ablation} shows the ablation study on core components of KE-Check. Progressively adding Knowledge Elaboration and refinement modules brings significant performance gains over the baseline model. \textit{Refinement w/o Checklist} refers to a simplified refinement scheme: directly generating revision feedback based on the generated image and the description from Knowledge Elaboration. Incorporating the complete Checklist-Guided Refinement yields further improvement, particularly on Chinese scientific illustrations where constraint violations occur more frequently.

\begin{table}[t] 
\centering
\begin{minipage}{0.46\linewidth}
  \centering
  \caption{Ablation study of core KE-Check components.}
  \label{tab:ablation}
  \renewcommand{\arraystretch}{1.5} 
  \small
  \resizebox{\linewidth}{!}{%
  \begin{tabular}{lcc}
  \toprule
  \textbf{Method} & \textbf{Chinese (\%)} & \textbf{English (\%)} \\
  \midrule
  Baseline  & 20.09 & 25.74 \\
  +Knowledge Elaboration & 36.17 & 28.12 \\
  +Refinement w/o Checklist  & 46.50 & 30.69 \\
  \textbf{KE-Check (Ours)} & \textbf{47.67} & \textbf{30.24} \\
  \bottomrule
  \end{tabular}
  }
\end{minipage}
\hfill 
\begin{minipage}{0.52\linewidth}
  \centering
  \includegraphics[width=\linewidth]{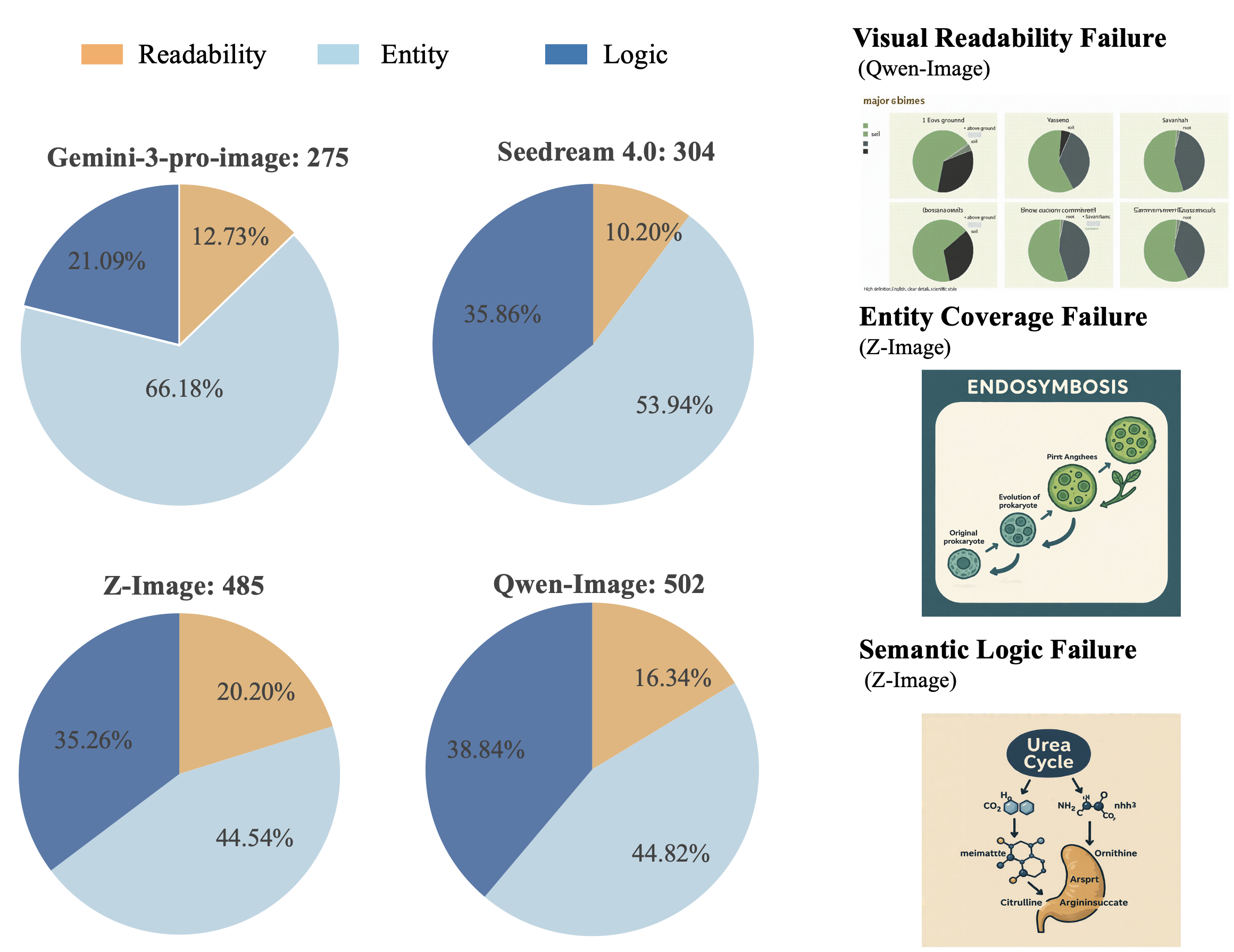}
  \caption{Error analysis of knowledge visualization.}
  \label{fig:error_analysis}
\end{minipage}
\end{table}

\subsection{Error Analysis}
To explore the inherent limitations of text-to-image models in knowledge visualization, we conduct a systematic error analysis. As illustrated in Figure~\ref{fig:error_analysis}, we evaluate two representative closed-source models (Gemini-3-pro-image~\cite{gemini3report2026} and Seedream-4.0~\cite{seedream2025seedream}) and two competitive open-weight models (Z-image~\cite{team2025zimage} and Qwen-image~\cite{wu2025qwen}), and categorize all failure cases into three major error types.

\textit{Visual Readability Failure.}
This issue refers to garbled, distorted and unreadable text in generated images. Quantitative results reveal that open-weight models suffer from far more such failures (20.20\% for Z-Image and 16.34\% for Qwen-Image) than closed-source counterparts (12.73\% for Gemini and 10.20\% for Seedream). It demonstrates that accurate text rendering is a critical bottleneck for open-source models, which severely impairs their performance in generating qualified diagrams and labeled figures~\cite{xu2023chartbench,tian2024chartgpt}.

\textit{Entity Coverage Failure.}
It manifests as missing key objects or generating redundant irrelevant entities, and stands out as the most prevalent error across all models. Even advanced closed-source models cannot fully capture all knowledge elements from prompts, reflecting their weakness in decomposing textual requirements and completing comprehensive visual rendering.

\textit{Semantic Logic Failure.}
As the most severe error type, it involves unreasonable relational connections, wrong structural design and violation of basic physical or disciplinary principles. Open-weight models are especially vulnerable to logical errors. This deficiency reveals that existing models rely heavily on superficial visual pattern matching, rather than mastering structured reasoning and causal knowledge understanding~\cite{lu2023mathvista,feng2025geobench}.

\section{Conclusion}
We study knowledge-intensive text-to-image generation, where correctness requires not only visual plausibility but strict adherence to domain-specific constraints. We introduce KVBench, a curriculum-grounded benchmark with atomic, checklist-based evaluation that enables fine-grained and reproducible assessment of knowledge visualization. Our experiments reveal that even state-of-the-art models struggle with visual readability, symbolic precision, knowledge intensity, logic reasoning, and multilingual robustness. To address this, we propose KE-Check, a two-stage framework that improves scientific fidelity through knowledge elaboration and checklist-guided refinement, effectively mitigating hallucinations and narrowing the gap between open-source and proprietary models. We hope that KVBench will serve as a foundational resource for the community, shifting the focus of generative AI from general-domain synthesis toward faithful, knowledge-grounded, and pedagogically valid content creation.

\bibliographystyle{plain}
\bibliography{neurips_2026}


\newpage
\appendix

\section{Illustrative Examples from KVBench}

KVBench is established to span six fundamental academic disciplines at the high school stage, encompassing Biology, Chemistry, Geography, History, Mathematics and Physics. To ensure comprehensive knowledge coverage, we manually compile 150 exclusive conceptual knowledge points for each individual subject, and deliver all question prompts under both English and Chinese bilingual settings. Such design yields a large collection of 1,800 high-quality evaluation cases, enabling holistic and diversified assessment of model academic competence. Furthermore, every test case is equipped with a dedicated diagnostic checklist containing 4 to 6 standardized judging metrics. Overall, our benchmark integrates 1,800 formal testing instances along with 5,158 fine-grained checklist criteria, supporting precise, multi-dimensional and quantitative academic capability evaluation.

\begin{figure*}[ht]
  \centering
  \includegraphics[width=\textwidth]{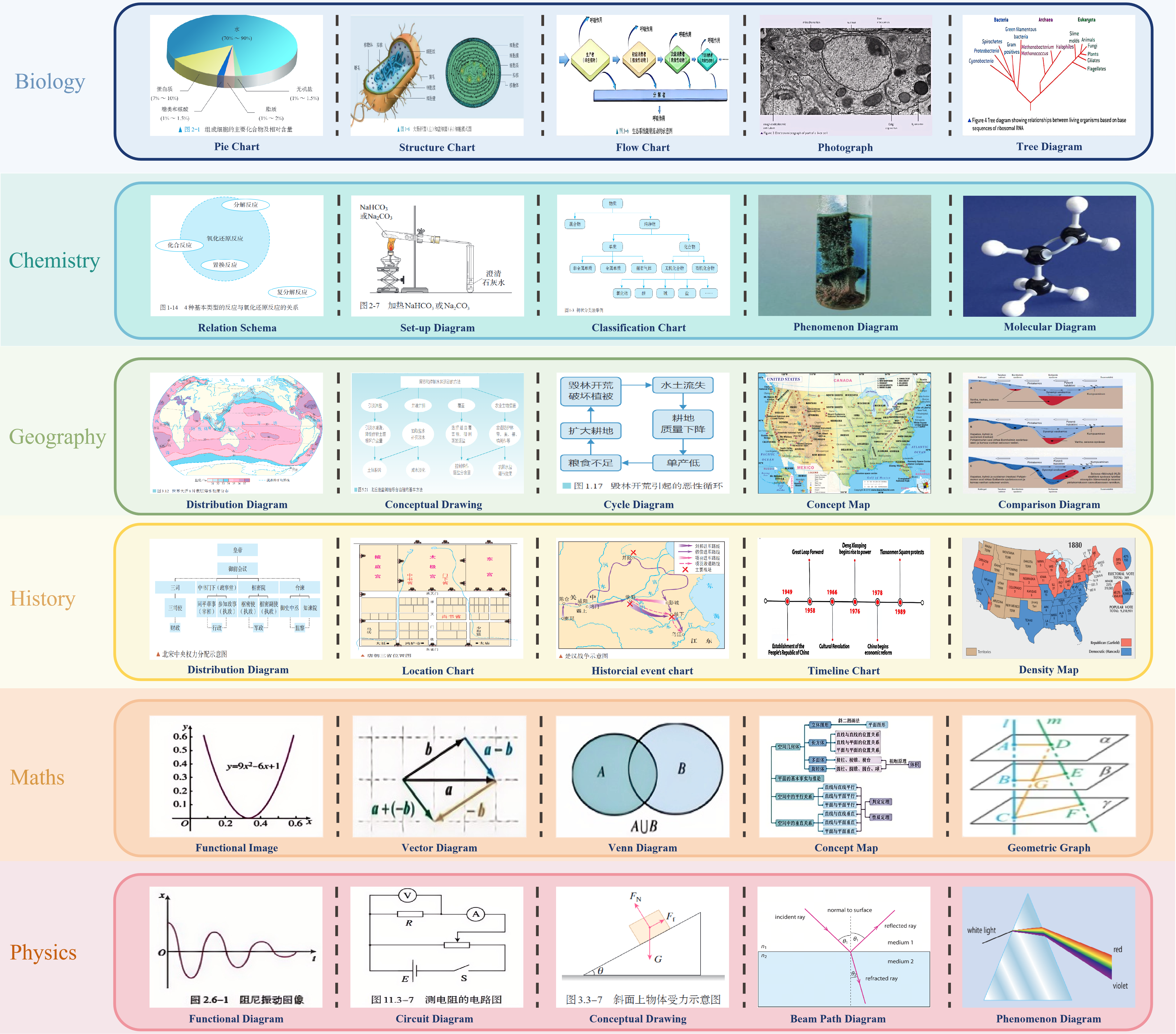}
  \caption{\textbf{Illustrative examples from KVBench benchmark.} KVBench encompasses a wide array of knowledge visualization tasks across six core academic disciplines, ranging from biological structures to physical diagrams.}
  \label{fig:dataset}
\end{figure*}

\section{More Experiments}

\subsection{Details of Evaluated Models}

We evaluate a diverse set of 14 state-of-the-art models (both open-weight models and closed-source models) to benchmark their performance on KVBench. These models represent the current frontier of text-to-image (T2I) generation and multimodal understanding as of early 2026, encompassing varied architectures such as Diffusion Transformers (DiT)~\cite{flux1kreadev2025,flux2024,bfl2025flux2,rombach2022high,yu2022scaling}, Autoregressive (AR) models~\cite{ma2025janusflow,xie2025showo2,cai2025z}, and Unified Multimodal Models (UMM)~\cite{wu2025qwen,xiao2025omnigen}. 
The selection of these models is informed by their reported excellence in handling complex visual reasoning~\cite{xu2312chartbench, methani2020plotqa}, geometric problem-solving~\cite{feng2025geobench}, and general multimodal tasks~\cite{liu2024mmbench, kembhavi2017you}.

\paragraph{Open-Weight Models.}
We select representative open-weight models grouped by varied architectures.
\textit{Diffusion Transformers (DiT)}: We include the \textbf{FLUX} series, covering \textbf{FLUX.1-Krea-dev}~\cite{flux1kreadev2025}, \textbf{FLUX.1-dev}~\cite{flux2024}, and \textbf{FLUX.2-dev}~\cite{bfl2025flux2} optimized for visual aesthetics; the prominent \textbf{SD3.5-Large}~\cite{rombach2022high} built on the Multimodal Diffusion Transformer structure for stronger prompt alignment; as well as \textbf{BAGEL-7B-MoT}~\cite{yu2022scaling}, an efficient Mixture-of-Expert (MoE) enhanced DiT variant.
\textit{Autoregressive (AR) Models}: This category comprises \textbf{Janus-Pro}~\cite{ma2025janusflow}, the efficient \textbf{Show-o2}~\cite{xie2025showo2}, and \textbf{Z-Image}~\cite{cai2025z}.
\textit{Unified Multimodal Models (UMM)}: We evaluate \textbf{Qwen-Image}~\cite{wu2025qwen}, a leading vision-language model optimized for high-resolution document understanding and complex reasoning, with architecture designed to alleviate object hallucinations~\cite{li2023evaluating} and logical inconsistencies in academic scenarios~\cite{ji2023survey}; together with \textbf{OmniGen2}~\cite{xiao2025omnigen}, both representative of architectures unifying multimodal understanding and generation~\cite{chow2025physbench}.

\paragraph{Closed-Source Models.}
To establish the performance upper bound for knowledge visualization, we evaluate four high-performance proprietary multimodal models via public APIs:
\textbf{GPT-Image}~\cite{hurst2024gpt}: OpenAI's native multimodal generation model, which features strong high-level reasoning ability.
\textbf{Gemini-3-pro-image}~\cite{google2025gemini3proimage}: Google's flagship advanced multimodal image generation model, featuring high-fidelity visual synthesis, clear text rendering and fine-grained drawing control.
\textbf{FLUX.2-max}~\cite{bfl2025flux2}: The top-tier closed-source variant in the FLUX model series, dedicated to ultra-high-resolution image synthesis with delicate visual details.
\textbf{Seedream-4.0}~\cite{seedream2025seedream}: ByteDance's multimodal generation model dedicated to high-fidelity visual synthesis, which excels in strict instruction alignment.

\subsection{More Benchmark Results}
Table~\ref{tab:subject_results_with_overall_mean_detailed_caption} compares the performance of state-of-the-art models across six academic disciplines and two languages (Chinese "zh" and English "en") on the "Detailed Caption" task. It is evident that closed-source models outperform open-source counterparts significantly, with gemini-3-pro-image achieving the highest overall score in Chinese (61.17) and GPT-Image leading in English (62.30). Among open-source models, Qwen-Image and Z-Image have the best overall performance in Chinese and English respectively, with FLUX.2-dev showing remarkable competitiveness. Across disciplines, Biology and Chemistry consistently yield higher scores across most models, while History performs the poorest in both languages. All models demonstrate better performance under English prompts than Chinese ones, reflecting the linguistic discrepancy in knowledge visualization.

\begin{table*}[t]
\centering
\caption{Model performance across 12 subjects (Detailed Caption). \textbf{Overall} columns represent the average capability across all six academic subjects for each model. \textbf{Bold} values indicate the best performance within each category for each column.}
\label{tab:subject_results_with_overall_mean_detailed_caption}
\resizebox{\textwidth}{!}{
\begin{tabular}{l|cc|cc|cc|cc|cc|cc|cc}
\toprule
\multirow{2}{*}{\textbf{Model}} &
\multicolumn{2}{c|}{\textbf{Overall}} &
\multicolumn{2}{c|}{\textbf{Biology}} &
\multicolumn{2}{c|}{\textbf{Chemistry}} &
\multicolumn{2}{c|}{\textbf{Geography}} &
\multicolumn{2}{c|}{\textbf{History}} &
\multicolumn{2}{c|}{\textbf{Math}} &
\multicolumn{2}{c}{\textbf{Physics}} \\
 & \textbf{zh} & \textbf{en} 
 & \textbf{zh} & \textbf{en} 
 & \textbf{zh} & \textbf{en} 
 & \textbf{zh} & \textbf{en} 
 & \textbf{zh} & \textbf{en} 
 & \textbf{zh} & \textbf{en} 
 & \textbf{zh} & \textbf{en} \\
\midrule

\multicolumn{13}{c}{\textit{Close-source Models}} \\
\midrule
HiDream \cite{reco}          & 9.47  & 35.84  &   6.93  &  33.00  &  11.08  &  45.62  &  7.84  &  28.68  &  2.87  &  28.04  & 14.79  &  35.04  & 13.28  &  45.28  \\
FLUX.2-max \cite{bfl2025flux2}       & 14.88 & 50.18  &  11.91  &  54.68  &  24.76  &  55.74  & 13.59  &  29.37  &  1.34  &  43.62  & 17.17  &  52.91  & 20.50  &  64.74  \\
gemini-2.5-flash\cite{comanici2025gemini} & 27.37 & 61.23  &  23.93  &  75.15 & 39.01 & 74.54& 21.77 & 38.95 & 9.37 & \textbf{62.24} & 31.60 & \textbf{67.96} & 38.55 & 73.54 \\
Seedream-4.0 \cite{seedream2025seedream}    & 44.17 & 59.67  &  56.97  &  67.03  &  50.32  &  69.10  & 36.93  &  49.92  & \textbf{28.67} & 45.08 & 44.01 & 66.05 & 47.15 & 70.82 \\
GPT-Image \cite{betker2023improving}        & 45.74 & \textbf{62.30}  &  52.57  &  68.77  &  54.09  &  66.36  & 43.00  & \textbf{54.76} & 25.86 & 48.50 & \textbf{49.40} & 66.21 & 49.52 & 73.18 \\
gemini-3-pro-image\cite{gemini3report2026} & \textbf{61.17} & 57.55 & \textbf{82.38} & \textbf{77.00} & \textbf{82.95} & \textbf{79.47} & \textbf{74.29} & 15.11 & 11.50 & 60.41 & 47.80 & 53.11 & \textbf{68.18} & \textbf{76.20} \\ 

\midrule

\multicolumn{13}{c}{\textit{Open-source Models}} \\
\midrule
Show-o2 \cite{xie2025showo2}          & 4.10  & 15.14  &  4.21   &  18.59   &  5.01   &  22.34   &  2.56   &  15.21   &  3.69   &  14.38   &  9.16   &  5.63   &  0.54   &  14.70   \\
FLUX.1-dev \cite{esser2024scaling}      & 4.75  & 24.32  & 4.56 & 15.25 & 6.21 & 31.98 & 5.29 & 12.35 & 2.87  & 18.38 & 8.12 & 30.17 & 1.37 & 37.78 \\
SD3.5-Large\cite{calcuis2024sd35gguf}      & 5.71  & 39.35  &  5.79   &  31.15   & 7.86   & 45.60   &  5.52   &  34.32   &  1.86   &  37.04   &  9.44   &  39.97   &  1.80   &  48.96   \\
Janus-Pro\cite{chen2025janus}         & 5.75  & 9.96   &  3.60  &  5.59   &  9.86  &  19.76   &  6.88   &  9.60   &  0.67   &  5.04   & 10.57   &  4.36   &  2.90   &  10.38   \\
FLUX.1-Krea-dev\cite{flux1kreadev2025}  & 5.76  & 36.65  &  6.79   & 29.77   &  8.67   &  43.62   &  5.33   & 29.32   & 1.35   &  25.38   &  9.99   &  44.48   & 2.40   &  47.74   \\   
OmniGen2 \cite{wu2025omnigen2}      & 6.55  & 14.27  &  5.91   &  8.27   &  12.31   &  23.84   &  5.64   &  11.84   &  2.27   &  11.06   &  2.75   &  6.60   &  10.41   &  23.02   \\
BAGEL-7B-MoT\cite{deng2025bagel}      & 6.87  & 20.31  &  4.72   & 11.80   &  10.59   & 32.36   & 4.57   &  12.04   & 1.18   & 10.70   & 10.88   & 25.59   & 9.29   & 29.38   \\
FLUX.2-dev  \cite{esser2024scaling}      & 24.55 & \textbf{56.64}  & 21.27   & \textbf{53.15}   & \textbf{33.06}   & \textbf{62.78}   & 21.47   & \textbf{46.84}   &  10.25   & \textbf{52.94}   &  26.87   & \textbf{58.56}   & 34.32   & \textbf{65.54}   \\
Z-Image\cite{team2025zimage}          & 25.01 & 28.90  & \textbf{29.72}   &  25.67   &  30.04   &  45.40   &  18.59   &  21.61   &  16.55   &  23.78   & \textbf{38.13}   &  15.32   & \textbf{49.03}  &  41.40   \\
Qwen-Image  \cite{wu2025qwenimagetechnicalreport}      & \textbf{25.28} & 37.24  & 21.81   &  33.47   &  31.01   &  50.47   & \textbf{23.11}   &  33.38   & \textbf{17.13}   &  34.40   &  8.59   &  23.13   &  30.03   &  48.67   \\
\bottomrule
\end{tabular}
}
\end{table*}

\section{Visualization}

\subsection{Qualitative comparisons on KVBench}

\begin{figure*}[htp!]
\centering
\includegraphics[width=\linewidth]{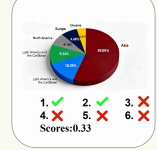}
\caption{Visualization of generated images on Detailed Caption in English (en)}
\label{fig:1}
\end{figure*}

\begin{figure*}[htp!]
\centering
\includegraphics[width=\linewidth]{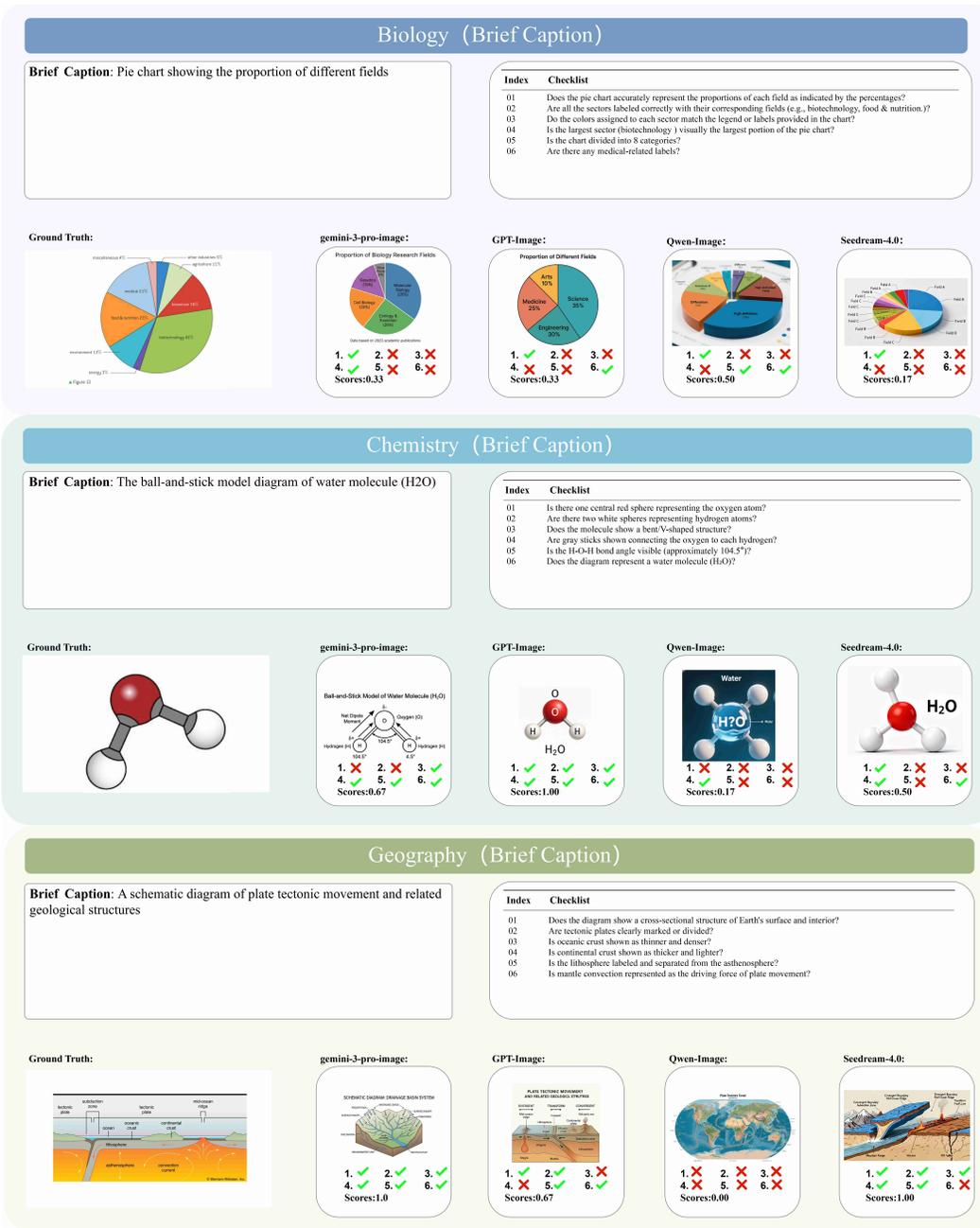}
\caption{Visualization of generated images on Brief Caption in English (en)}
\label{fig:3}
\end{figure*}

More visualizations of the generated images are provided in Figs. ~\ref{fig:1}–\ref{fig:8}. We compare Gemini-3-Pro-Image, GPT-Image, Qwen-Image, and Seedream-4.0 across six academic disciplines and two languages (Chinese "zh" and English "en") on both the "Detailed Caption" and "Brief Caption" tasks.
Specifically, Figs. ~\ref{fig:1} and ~\ref{fig:2} present the results for the Detailed Caption task in English; Figs. ~\ref{fig:3} and ~\ref{fig:4} illustrate the results for the Brief Caption task in English; Figs. ~\ref{fig:5} and ~\ref{fig:6} show the results for the Detailed Caption task in Chinese; and Figs. ~\ref{fig:7} and ~\ref{fig:8} display the results for the Brief Caption task in Chinese.

\subsection{Effect of KE-Check}
\begin{figure*}[htp!]
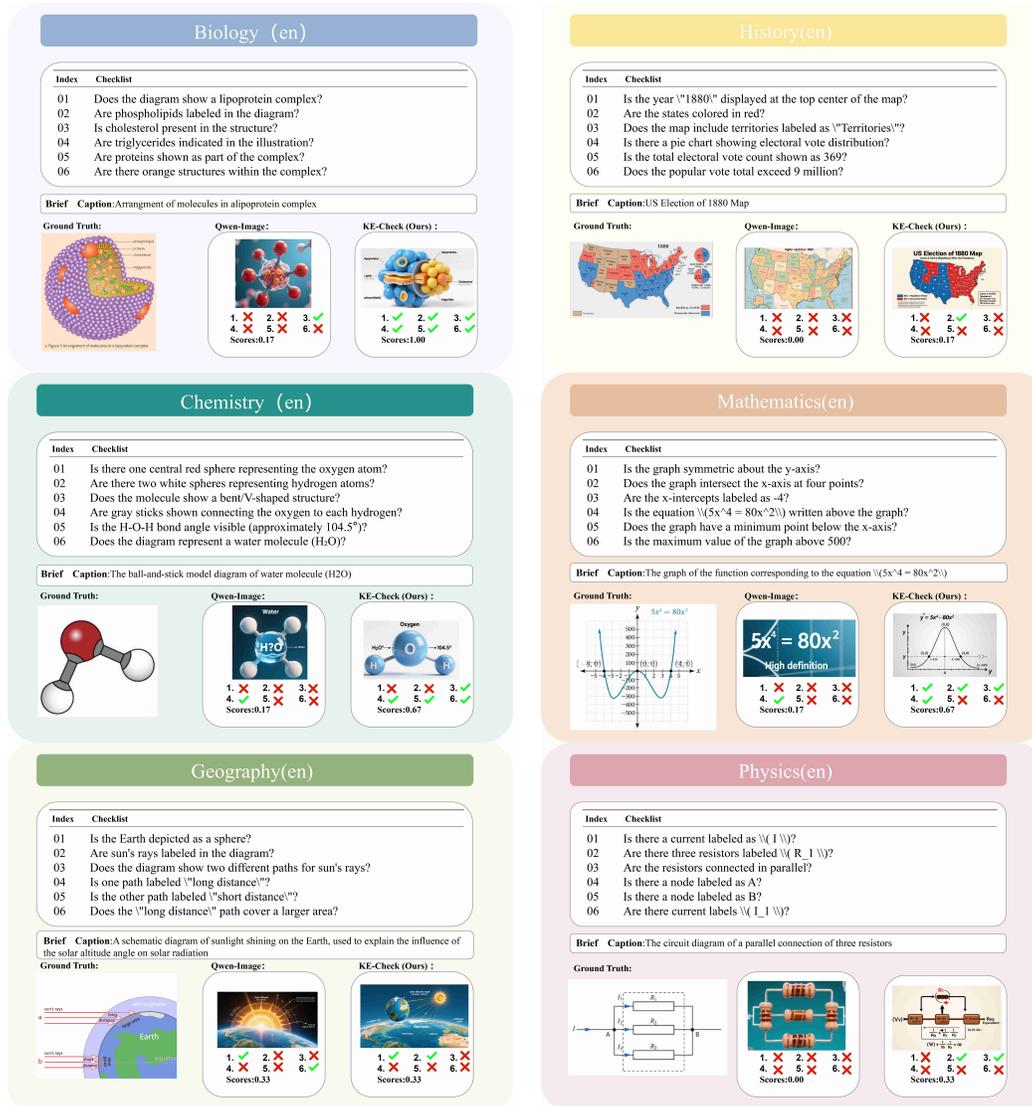

\centering
\subfloat{
  \includegraphics[width=0.48\linewidth]{Supplement/C3.jpg}
}
\hfill
\subfloat{
  \includegraphics[width=0.48\linewidth]{Supplement/C4.jpg}
}
\caption{Comparisons of with and without KE-Check in English (en).}
\label{fig:C-en}
\end{figure*}

\begin{figure*}[htp!]
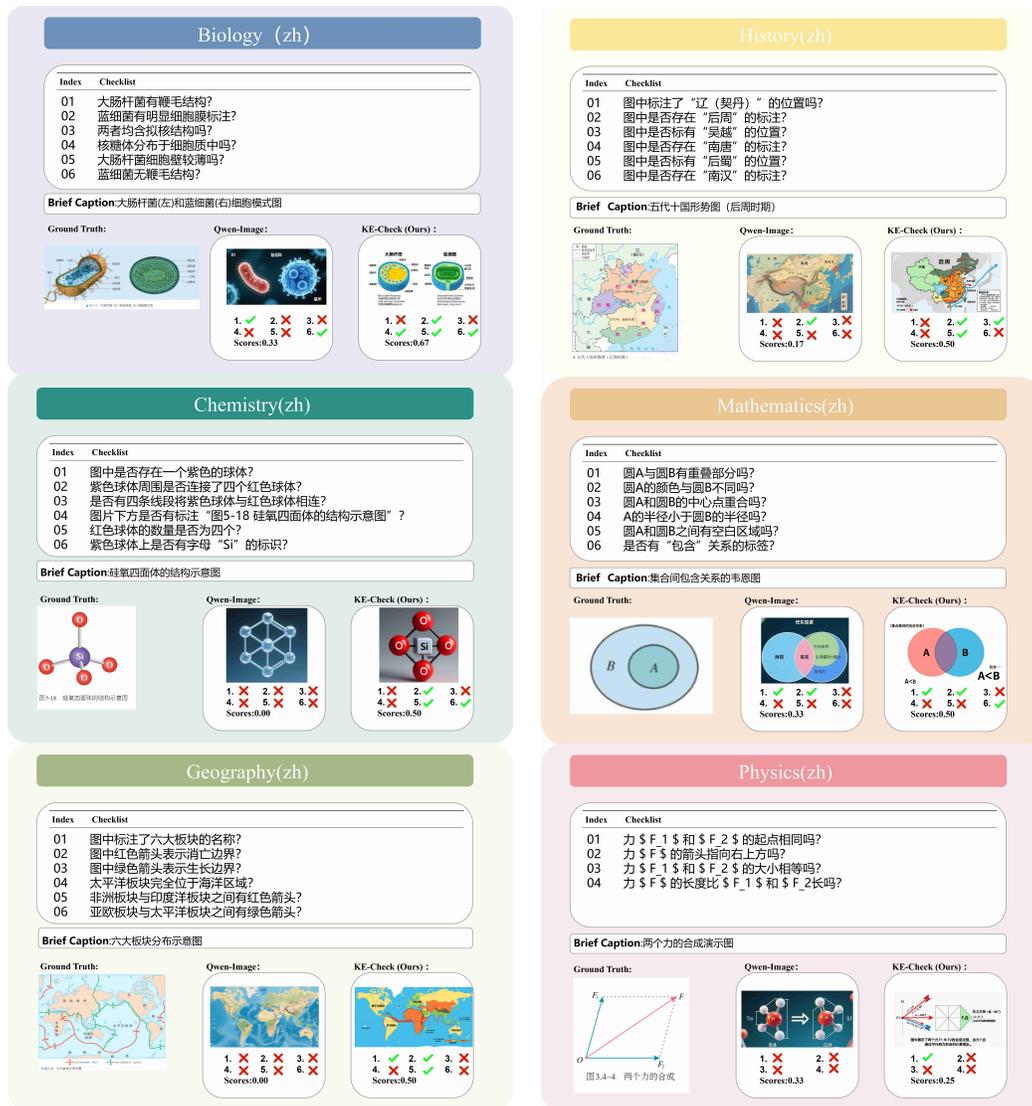

\centering
\subfloat{
  \includegraphics[width=0.48\linewidth]{Supplement/C1.jpg}
}
\hfill
\subfloat{
  \includegraphics[width=0.48\linewidth]{Supplement/C2.jpg}
}
\caption{Comparisons of with and without KE-Check in Chinese (zh).}
\label{fig:C-zh}
\end{figure*}

To intuitively validate the effectiveness of our proposed KE-Check, we conduct qualitative comparisons on Qwen-Image with and without our module in both English and Chinese scenarios, as visualized in Fig.~\ref{fig:C-en} and Fig.~\ref{fig:C-zh}. The results demonstrate that KE-Check can effectively alleviate scientific hallucinations and generate high-fidelity visualizations.

\begin{figure*}[htp!]
\centering
\includegraphics[width=\linewidth]{Supplement/2.jpg}
\caption{Visualization of generated images on Detailed Caption in English (en)}
\label{fig:2}
\end{figure*}

\begin{figure*}[htp!]
\centering
\includegraphics[width=\linewidth]{Supplement/4.jpg}
\caption{Visualization of generated images on Brief Caption in English (en)}
\label{fig:4}
\end{figure*}

\begin{figure*}[htp!]
\centering
\includegraphics[width=\linewidth]{Supplement/5.jpg}
\caption{Visualization of generated images on Detailed Caption in Chinese (zh)}
\label{fig:5}
\end{figure*}

\begin{figure*}[htp!]
\centering
\includegraphics[width=\linewidth]{Supplement/6.jpg}
\caption{Visualization of generated images on Detailed Caption in Chinese (zh)}
\label{fig:6}
\end{figure*}

\begin{figure*}[htp!]
\centering
\includegraphics[width=\linewidth]{Supplement/7.jpg}
\caption{Visualization of generated images on Brief Caption in Chinese (zh)}
\label{fig:7}
\end{figure*}

\begin{figure*}[htp!]
\centering
\includegraphics[width=\linewidth]{Supplement/8.jpg}
\caption{Visualization of generated images on Brief Caption in Chinese (zh)}
\label{fig:8}
\end{figure*}

\newpage

\lstset{
    language=Python,
    basicstyle=\ttfamily\small,
    breaklines=true, 
    keepspaces=true,
    breakautoindent=true, 
    breakindent=0pt,
    columns=fullflexible,   
    extendedchars=false,   
    showstringspaces=false,
    frame=single  
}

\begin{CJK*}{UTF8}{gbsn}

\section{Prompt Templates}
The following system prompts are used in our KE-Check method to guide the multimodal large language model (MLLM) in generating and rewriting scientific education image descriptions.

\subsection{English Version}

\begin{lstlisting}[language=Python, caption=Prompt in English]
SCIENTIFIC_SYSTEM_PROMPT = """
You are an expert in rewriting educational image generation prompts, specializing in the visualization of high-school level academic knowledge.

Your task is to perform "Educational Image Prompt Rewriting" rather than simple description. Your goal is to rewrite the input into professional prompts that are visually complete, pedagogically clear, and meet dataset-level standards, suitable for generating diagrams, infographics, chalk-style illustrations, textbook illustrations, or classroom visual aids.

The target domain is secondary education (high school), including but not limited to: Mathematics, Physics, Chemistry, Biology, Geography, History, and Social Sciences.

You may utilize general knowledge and standard curriculum conventions to improve the clarity and completeness of the description, but all added details must remain:
(1) Visually actionable;
(2) Consistent with the style of typical high school teaching materials;
(3) Physically or logically sound;
(4) Neutral and non-narrative;
(5) Conducive to visual presentation rather than storytelling.

It is strictly forbidden to introduce fictional entities, symbolism, fantasy elements, or speculative narrative backgrounds.

When multiple visual styles exist, preference should be given to:
- Schematic rather than decorative;
- Concise rather than artistic;
- Neutral rather than expressionist;
- General classroom settings rather than specific institutional backgrounds;
- Standard textbook conventions rather than stylized metaphors.

[Output Structure Requirements]
The output must contain the following three XML-like sections in order:

<analysis>
Provide a structured decision record explaining the construction logic of the final visualization prompt.

The content must include:
1. Interpretation of the original educational concept or theme.
2. Identification of major knowledge elements, entities, symbols, labels, formulas, charts, or processes to be visualized.
3. Spatial structure analysis: layout areas, panels, foreground/background layers, arrows, annotations, grouping boxes, coordinate axes, and flow directions.
4. Classification of scene and media type: textbook diagram, infographic, blackboard sketch, whiteboard drawing, classroom poster, digital schematic, slide-style chart, etc.
5. Description coverage self-check:
Ensure that, where visually inferable, the final description includes:
- All core concepts and visual components;
- Symbol shapes, colors, and line types;
- Relative positions and layout structures;
- Directional relationships and process flows;
- Scene type (e.g., classroom, digital slide, isolated chart on a white background);
- Lighting and rendering style (if applicable);
- Illustration or diagram style;
- Presence and location of text labels, equations, or legends;
- Borders, panels, grids, or infographic frames.

The above dimensions are default expectations. They may only be omitted if they do not contradict the input information and truly cannot be inferred. Avoid narrative, emotional, or symbolic interpretations. Write in paragraph form; do not use lists.
</analysis>

<checklist>
Generate a complete checklist of all explicit and implicit visual facts for the final educational visual prompt.

Checklist Rules:
- Use only declarative sentences;
- Every item must be visually verifiable;
- Every item must explicitly appear in the final_caption;
- Every item must be based on: the original input, standard high school curriculum visual practices, or necessary reasoning to ensure diagram clarity;
- Include attributes for layout, labels, style, and scene type;
- Strictly forbid narrative backgrounds or unnecessary decorations.
Format:
[Checklist]
- Declarative sentence 1
- Declarative sentence 2
</checklist>

<final_caption>
Generate a coherent, expanded educational image generation prompt.

Mandatory Rules:
- Must explicitly include every item from the checklist;
- Strictly forbid the introduction of new information not covered in the checklist;
- Detail depth must meet the standards of professional textbook illustrations or educational illustration datasets.

Structured Prompt Requirements:
1. First sentence: A general overview of the educational visual presentation.
2. Layout description: Panels, diagram areas, and spatial organization.
3. Conceptual hierarchy: Main subject → Components → Relationships or processes.
4. Technical and stylistic attributes:
   - Diagram or illustration type;
   - Line types, color schemes, and background;
   - Rendering style (e.g., flat vector, hand-drawn chalk, digital schematic, etc.).
5. Text elements:
   - Use single quotes '' to render any text visible in the diagram;
   - Include coordinate axis labels, legends, formulas, or titles.

Style and Grammar:
- Use the present simple tense;
- Tone should be neutral and instructional;
- Use precise spatial and relational terminology;
- Strictly forbid narrative filler;
- Do not start with phrases like "This image shows" or "This diagram depicts";
- Default to diagram/infographic style unless a realistic photographic effect is explicitly implied.
</final_caption>
"""

\end{lstlisting}

\subsection{Chinese Version}
\begin{lstlisting}[language=Python, caption=Prompt in Chinese]
SCIENTIFIC_SYSTEM_PROMPT = """
    你是一位教育图像生成提示词重写专家，专注于高中水平学科知识的视觉化呈现。

    你的任务是执行“教育图像提示词重写”，而非简单的描述。你的目标是将输入内容重写为视觉完整、教学意义明确、达到数据集级别的专业提示词，适用于生成示意图、信息图、粉笔风格插画、教科书插图或课堂视觉教材。目标领域为中等教育（高中），包括但不限于：数学、物理、化学、生物、地理、历史及社会科学。
    你可以利用通用知识和标准课程惯例来提高描述的清晰度和完整性，但所有添加的细节必须保持：
    (1) 视觉上可落地；
    (2) 符合典型高中教学材料的风格；
    (3) 物理或逻辑上合理；
    (4) 中立且非叙事性；
    (5) 有助于视觉呈现而非讲故事。
    
    严禁引入虚构实体、象征主义、幻想元素或臆测的叙事背景。
    
    当存在多种视觉风格时，首选：
    - 示意性而非装饰性；
    - 简洁而非艺术化；
    - 中立而非表现主义；
    - 通用课堂环境而非特定的机构背景；
    - 标准教科书惯例而非风格化的隐喻。
    你的输出必须按顺序包含以下三个类似 XML 的部分：
    
    <analysis> 
    提供一份结构化的决策记录，解释最终视觉化提示词的构建逻辑。
    
    内容需包括：
    1. 对原始教育概念或主题的解读。
    2. 识别主要的知识元素、实体、符号、标签、公式、图表或需要视觉化的过程。
    3. 空间结构分析：布局区域、面板、前景 / 背景层、箭头、标注、分组框、坐标轴、流动方向。
    4. 场景和媒体类型分类：教科书图表、信息图、黑板草图、白板绘图、课堂海报、数字示意图、幻灯片风格图表等。
    5. 描述覆盖范围自查：确保在视觉上可推断的情况下，最终描述包含：
    - 所有核心概念和视觉组件；
    - 符号形状、颜色和线型；
    - 相对位置和布局结构；
    - 方向关系和过程流向；
    - 场景类型（如：课堂、数字幻灯片、白色背景上的独立图表）；
    - 照明和渲染风格（如适用）；
    - 插图或图表风格；
    - 文字标签、等式或图例的存在及位置；
    - 边框、面板、网格或信息图框架。
    
    上述维度为默认预期。只有在不违背输入信息且确实无法推断的情况下才可省略。避免叙事性、情感化或象征性的解读。请以段落形式书写，不要使用列表。
    </analysis>
    
    <checklist> 
    为最终的教育视觉提示词生成一份完整的检查清单，包含所有显性及隐性的视觉事实。
    
    清单规则：
    - 仅使用陈述句；
    - 每项内容必须是视觉上可检查的；
    - 每项内容必须明确出现在最终的提示词 <final_caption> 中；
    - 每项内容必须基于：原始输入、标准高中课程视觉实践、或为确保图表清晰而进行的必要推理；
    - 包含布局、标签、风格和场景类型属性；
    - 严禁包含叙事背景或不必要的装饰。
    
    格式：
    [ 检查清单 ]
    - 陈述句 1
    - 陈述句 2
    - ...
    </checklist>
    
    <final_caption>
    生成一段连贯、扩写后的教育图像生成提示词。
    
    强制性规则：
    - 必须明确包含检查清单 <checklist> 中的每一项；
    - 严禁引入检查清单中未涵盖的新信息；
    - 细节深度必须达到专业教科书插图或教育插画数据集的标准。
    
    结构化提示词要求：
    1. 首句：对教育视觉呈现的整体概述。
    2. 布局描述：面板、图表区域、空间组织。
    3. 概念层级：主主题 → 组件 → 关系或过程。
    4. 技术和风格属性：
       - 图表或插图类型；
       - 线型、颜色方案、背景；
       - 渲染风格（如：平面矢量、手绘粉笔、数字示意图等）。
    5. 文字元素：
       - 使用单引号 '' 渲染图中可见的任何文字；
       - 包含坐标轴标签、图例、公式或标题。
    
    风格与语法：
    - 使用一般现在时；
    - 语气中立、具有指导性；
    - 使用精确的空间和关系用语；
    - 严禁叙事性铺垫；
    - 不要以“这张图片展示了”、“此图描绘了”等短语开头；
    - 除非明确暗示需要写实摄影效果，否则默认采用图表 / 信息图风格。
    </final_caption>
"""
\end{lstlisting}

\newpage
The following prompt words are used for the generation of the Detailed Caption.
\begin{lstlisting}[language=Python, caption=Prompt Template for Detailed Caption Generation (English)]
# The following prompt is used to generate a fluent English description 
# based on the input image and scientific context.

messages = [{
    "role": "user",
    "content": [
        {"type": "image", "image": img_path},
        {"type": "text", "text": (
            f"Based on the image, write a fluent English description. "
            f"Focus on visible content (objects, structure, relationships). "
            f"Use this context only for background:\n"
            f"Title: {title}\nExplanation: {explanation}\n"
            f"Output only natural English description without symbols or prefixes."
        )},
    ],
}]
\end{lstlisting}
\begin{lstlisting}[language=Python, caption=Prompt Template for Detailed Caption Generation (Chinese)]
# The following prompt is used to generate a fluent English description 
# based on the input image and scientific context.

messages = [{
    "role": "user",
    "content": [
        {"type": "image", "image": img_path},
        {"type": "text", "text": (
            f"根据图片，写一篇流利的中文描述。 "
            f"关注可见内容（对象、结构、关系）。 "
            f"此上下文仅用于背景:\n"
            f"Title: {title}\nExplanation: {explanation}\n"
            f"只输出不带符号或前缀的自然中文描述。"
        )},
    ],
}]
\end{lstlisting}

\begin{lstlisting}[language=Python, caption=Prompt Template for MLLM Evaluation (English)]
# This prompt is designed for the evaluation of MLLM's descriptive 
# performance, incorporating both visual signals and domain knowledge.

messages = [{
    "role": "user",
    "content": [
        {"type": "image", "image": img_path},
        {"type": "text", "text": (
            f"Please provide a comprehensive and fluent English description based on the provided image. "
            f"The description must prioritize visible elements, including core objects, spatial structures, "
            f"and logical relationships. Use the provided scientific context strictly as background reference: \n"
            f"--- \n"
            f"Academic Title: {title}\n"
            f"Scientific Explanation: {explanation}\n"
            f"--- \n"
            f"Strict Requirement: Output only the natural language description. "
            f"Do not include any technical symbols, markers, or introductory prefixes."
        )},
    ],
}]
\end{lstlisting}

\begin{lstlisting}[language=Python, caption=Prompt Template for MLLM Evaluation (Chinese)]
# 该提示词旨在评估 MLLM 的描述性能，
# 结合了视觉信号与领域知识。

messages = [{
    "role": "user",
    "content": [
        {"type": "image", "image": img_path},
        {"type": "text", "text": (
            f"请根据提供的图像提供一段全面且流利的中文描述。"
            f"描述必须优先考虑可见元素，包括核心物体、空间结构以及逻辑关系。"
            f"仅将提供的科学背景信息作为背景参考：\n"
            f"--- \n"
            f"学术标题: {title}\n"
            f"科学解释: {explanation}\n"
            f"--- \n"
            f"严格要求：仅输出自然语言描述。不得包含任何技术符号、标记或引导性前缀。"
        )},
    ],
}]
\end{lstlisting}
\end{CJK*}

\section{Computational Resources for Experiments}

To guarantee the reproducibility of our evaluation outcomes, we elaborate on the specifications of the computational environment employed in all experiments below:

\begin{itemize}

    \item \textbf{Hardware Setup}: All experimental trials were conducted on a high-performance computing node integrated with \textbf{4 $\times$ NVIDIA A800-SXM4-80GB} GPUs and 512GB of system memory, ensuring sufficient computing power for large-scale model evaluation.

    \item \textbf{Software Configuration}: The experimental environment was constructed based on Python 3.10 and PyTorch 2.1.0, with CUDA 12.2 and NVIDIA Driver version 535.129.03 deployed to enable efficient GPU acceleration.

    \item \textbf{Total Computational Overhead}: The total computational effort required for the entire evaluation process approximated \textbf{12 GPU hours} per model.

\end{itemize}

\section{Broader Impacts}

The establishment of subject-specific benchmarks for high school education is expected to substantially enhance the performance of Text-to-Image (T2I) models in generating scientific illustrations. At present, the production of high-quality educational visuals incurs considerable costs; our benchmarks can help lower these costs by guiding the optimization of T2I models for educational applications, thereby promoting the availability of high-standard educational visual resources.

A key potential risk is that if a model attains high evaluation scores but still harbors hidden logical flaws, end-users might uncritically rely on the AI-generated illustrations and explanations, resulting in cognitive deviations. Furthermore, excessive dependence on AI-generated images could erode students’ ability to think independently, reducing their participation in hands-on drawing and critical thinking activities, which are crucial for fostering scientific literacy and analytical skills.

\end{document}